\documentclass[12pt]{article}
\usepackage[top=1in, bottom=1in, left=1in, right=1in]{geometry}

\usepackage{cite}
\usepackage{amsmath,amssymb,amsfonts}
\usepackage{algorithmic}
\usepackage{graphicx}
\usepackage{textcomp}
\usepackage{xcolor}
\usepackage{adjustbox}
\usepackage{booktabs}
\usepackage{multirow}
\usepackage{subcaption}
\usepackage{pgfplots}
\usepackage[absolute]{textpos}
\usepackage{float}
\usepackage{fancyhdr}

\begin{document}

\vspace{-1.5em}
\begin{flushright}
\end{flushright}

\begin{center}
    \Large \textbf{Knowledge Distillation Approach for SOS Fusion Staging: \\ Towards Fully Automated Skeletal Maturity Assessment}
    
    \vspace{1em}
    \normalsize
    Omid Halimi Milani\textsuperscript{1}, Amanda Nikho\textsuperscript{2}, Marouane Tliba\textsuperscript{1,3}, Lauren Mills\textsuperscript{2}, \\
    Ahmet Enis Cetin\textsuperscript{1}, Mohammed H Elnagar\textsuperscript{2} \\
    
    \vspace{0.5em}
    \textsuperscript{1}Department of Electrical and Computer Engineering, University of Illinois Chicago, USA \\
    \textsuperscript{2}Department of Orthodontics, College of Dentistry, University of Illinois Chicago, USA \\
    \textsuperscript{3}University of Orleans, France \\
\end{center}
\begin{abstract}
We introduce a novel deep learning framework for the automated staging of spheno-occipital synchondrosis (SOS) fusion, a critical diagnostic marker in both orthodontics and forensic anthropology. Our approach leverages a dual-model architecture wherein a teacher model, trained on manually cropped images, transfers its precise spatial understanding to a student model that operates on full, \textbf{uncropped images}. This knowledge distillation is facilitated by a newly formulated loss function that aligns spatial logits as well as incorporates \textbf{gradient-based attention spatial mapping}, ensuring that the student model internalizes the anatomically relevant features without relying on external cropping or YOLO-based segmentation. By leveraging expert-curated data and feedback at each step, our framework attains robust diagnostic accuracy, culminating in a clinically viable end-to-end pipeline. This streamlined approach obviates the need for additional pre-processing tools and accelerates deployment, thereby enhancing both the efficiency and consistency of skeletal maturation assessment in diverse clinical settings. 

\end{abstract}

\section{Introduction}
Accurate skeletal maturation assessment is crucial in both orthodontics and forensic anthropology. The spheno-occipital synchondrosis (SOS), located in the cranial base, is the final cartilaginous joint to fuse and thus serves as an important developmental indicator \cite{powell1963closure}. Its fusion stage informs key clinical decisions, including orthodontic treatment planning \cite{al-gumaei2022comparison, al-gumaei2023comparison}, estimation of pubertal growth spurts \cite{alhazmi2017timing}, and forensic age determination \cite{shirley2011spheno}.

In orthodontics, precise knowledge of skeletal maturation is paramount for guiding orthopedic jaw modifications. Growth discrepancies between the maxilla and mandible can often be mitigated if identified early, making timely assessment of maturation critical. Conventionally, skeletal age is evaluated via the cervical vertebrae method (CVM) on cephalograms or midsagittal slices from cone-beam computed tomography (CBCT) scans \cite{baccetti2005cervical,hassel1995skeletal}. However, when cervical vertebrae are unavailable or not clearly rendered, SOS fusion offers a valuable alternative indicator.

The SOS, as a synchondrosis, retains growth potential until late adolescence, which correlates with critical orthodontic interventions \cite{powell1963closure, alhazmi2021correlation, fernandez2016spheno}. For instance, evaluating the stage of SOS fusion helps determine the feasibility of rapid maxillary expansion (RME) \cite{tashayyodi2023relationship, bishara1987maxillary}, predict the likelihood of cranial-base-angle changes \cite{leonardi2021three}, and aid in timing intermaxillary procedures \cite{al-gumaei2022comparison,al-gumaei2023comparison}. Moreover, the onset of SOS fusion typically coincides with the pubertal growth spurt, reinforcing its utility as a surrogate for skeletal maturity \cite{alhazmi2017timing}. Beyond orthodontics, SOS fusion staging is also employed in forensic anthropology for estimating chronological age in young adults \cite{shirley2011spheno}.

Despite its clinical importance, manual SOS fusion staging is subject to inconsistencies stemming from varied classification schemes (ranging from three to six stages) \cite{alhazmi2021correlation, fernandez2016spheno, lottering2015ontogeny}, differing imaging modalities (e.g., histological sections, 2D radiography, CT, CBCT) \cite{krishan2013evaluation}, and observer variability. Such lack of standardization complicates the reproducibility and reliability of assessments in both research and clinical practice. Deep learning (DL) has shown growing potential in both dental maturity estimation~\cite{milani2024fully} and clinical orthodontic decision-making~\cite{rhee2025integrating}, motivating the development of more integrated and interpretable AI frameworks in craniofacial imaging.

Recent advances in convolutional neural networks (CNNs) demonstrate considerable promise for medical image analysis, particularly in noisy or detail-rich scenarios \cite{chen2024}. However, deploying a CNN for SOS staging faces unique challenges due to the subtle nature of the synchondrosis and the variability introduced by different imaging protocols. To address these issues, we introduce a DL framework that harnesses knowledge distillation to achieve accurate, automatic SOS fusion staging from CBCT scans.

Our method first trains a \emph{teacher} model using manually cropped SOS images generated under the direct supervision of a domain expert, ensuring highly precise local feature learning. The expert's real-time input during training helped refine both the cropping procedure and the identification of salient anatomical structures. Subsequently, these spatial representations are transferred to a \emph{student} model that processes full, uncropped CBCT images, thereby eliminating the need for external object detection models (e.g., YOLO). This transfer is realized through two key innovations: (1)~a \emph{spatial logits alignment loss} to align the student model's feature activations with those of the expert-trained teacher model; and (2)~a \emph{gradient-based attention loss} inspired by Grad-CAM \cite{selvaraju2017grad}, guiding the student network to anatomically pertinent areas.
Unlike methods requiring manual cropping or specialised region-of-interest localization, our framework processes complete images in an end-to-end fashion, reducing system complexity and potential segmentation errors. Our primary contributions can be summarized as follows: \begin{itemize} \item \textbf{A knowledge-distillation-based DL framework} that automates SOS fusion classification from CBCT images. \item \textbf{A spatial logits alignment loss}, enabling the student model to inherit high-fidelity spatial cues from the teacher model trained with expert-validated, manually cropped data. \item \textbf{A gradient-based attention loss}, ensuring robust focus on relevant anatomical structures without external segmentation models. \item \textbf{An end-to-end pipeline}, obviating the need for labor-intensive cropping or detection steps, thus enhancing clinical applicability. \end{itemize}

The rest of the paper is organized as follows: \textbf{Section~\ref{sec:dataset}} describes the dataset and manual segmentation procedure for the SOS region. \textbf{Section~\ref{sec:method}} details the proposed deep learning architecture and the knowledge distillation strategy. Finally, \textbf{Section~\ref{sec:discussion}} discusses experimental results, implications, and directions for future work.

\section{Dataset and Preprocessing for SOS Fusion Staging} 
\label{sec:dataset}


{\bf Patient Selection Criteria and Ethics Approval}

This IRB-exempt, retrospective study used de-identified CBCT scans (Study ID: STUDY2022-1048). Scans included patients (ages 7–76) in natural head position with relaxed facial muscles and extended  field-of-view (FOV). Exclusion criteria were craniofacial syndromes, trauma, or prior head/neck surgery. After removing ambiguous cases, the final dataset consisted of 723 scans (260 males, 370 females, 93 unknown; age 7–68).

{\bf SOS Fusion Staging}
The SOS fusion was classified into five stages as outlined by Bassed et al. (stages 1-5) \cite{bassed2010analysis}, with fusion beginning at the endocranial border and progressing inferiorly to the ectocranial border. In stage 1, the synchondrosis is completely open. In stage 2, the endocranial border has fused and the rest remains open. In stage 3, the ossification is halfway complete. By stage 4, fusion is complete throughout but still visible by means of a radiopaque scar. In stage 5, the scar is obliterated with even trabecular bone throughout. Classification was completed by three evaluators after robust training, including two orthodontists and an oral and maxillofacial radiologist. To investigate inter-rater reliability, a cross tabulation was computed using 97 scans. Cronbach’s alpha assesses the test-retest reliability between two investigators, and it was found to be 0.945, indicating that the inter reliability ($>0.80$) is very good.

{\bf Region of Interest (ROI) Selection and Manual Segmentation}
The CBCT scans were imported into Dolphin Imaging (version 11.95) using Digital Imaging and Communications in Medicine formation. Orientation proceeded in three planes using the 4-Equal-Slices-Volume-Layout view. In the coronal view, the axial plane was set parallel to the inferior border of the basilar part of the occipital bone by tilting the head; then, the axial plane was scrolled to its mid-vertical. In the axial view, the coronal plane was set parallel to the anterior border of the basilar part of the occipital bone by rotating the head; then, the coronal plane was scrolled to its mid-depth. In both the coronal and axial views, the sagittal plane was scrolled to evenly bisect the basilar part of the occipital bone. The data distribution across SOS Fusion classification stages includes 159 scans in stage 1, 92 in stage 2, 92 in stage 3, 125 in stage 4, and 255 in stage 5, totaling 723 scans.


Following, the Microsoft Snipping Tool (Version 11) was used to screen capture the entire midsagittal slice. The image was saved as a PNG file and staged by the evaluators. Using Microsoft Photos, the SOS was rotated until the ectocranial and endocranial ends aligned horizontally, and the synchondrosis was cropped to a standardized dimension of 2:1. The final dataset included 158 scans in stage 1, 88 in stage 2, 92 in stage 3, 124 in stage 4, and 252 in stage 5. See Fig.~\ref{fig:Orientation_and_Crop} for visualization of orientation and segmentation.
\begin{figure}[htbp]
\centering
\includegraphics[width=0.5\textwidth]{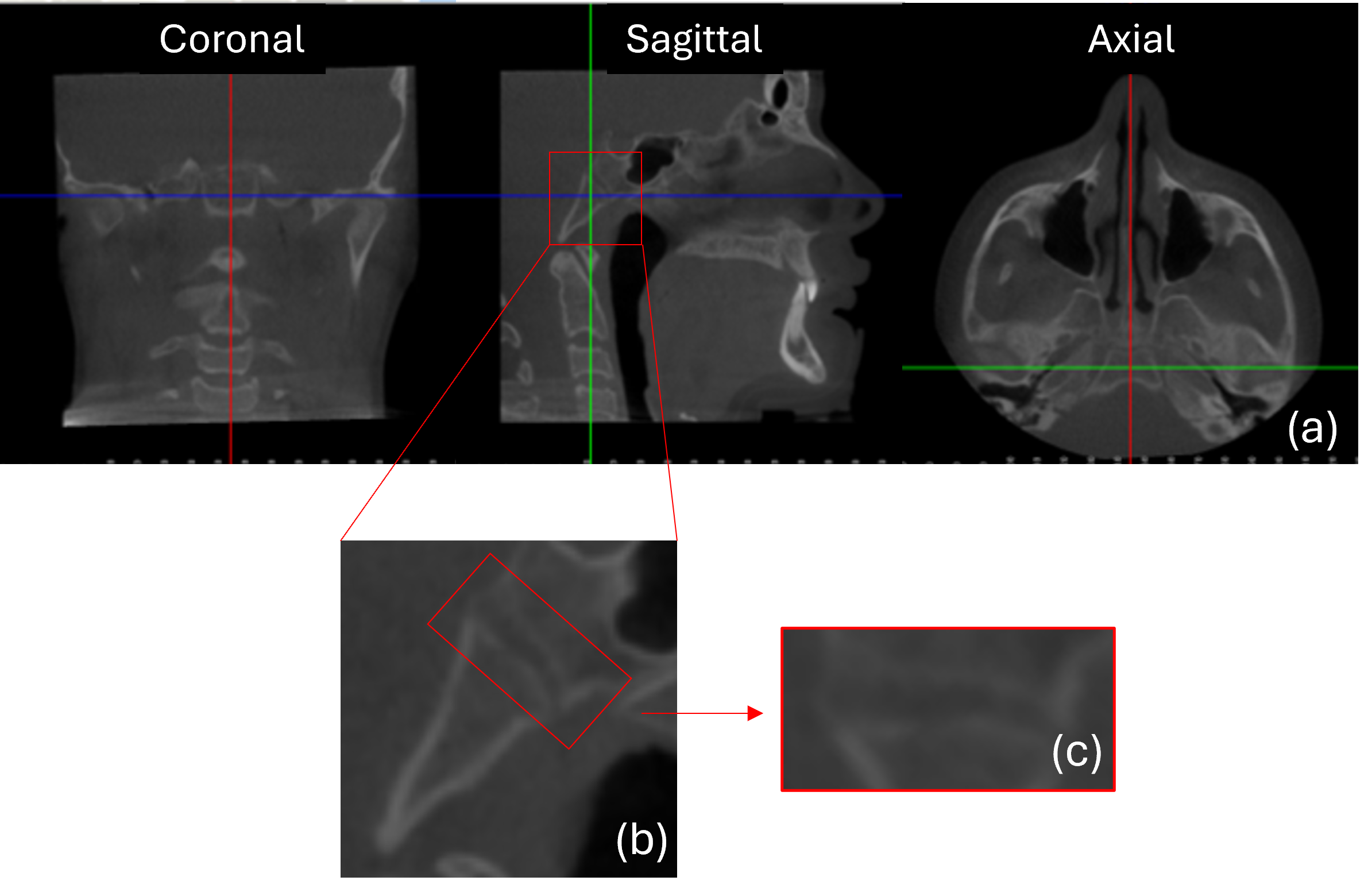}
\caption{Skull oriented in three planes (a). Occipital and
sphenoid bones cropped (b). SOS rotated and segmented (c).}
\label{fig:Orientation_and_Crop}
\end{figure}

\section{Enhancing Pre-Trained Models for Classification through Knowledge Distillation and Gradient-Based Attention}
\label{sec:method}

After curating and preprocessing the dataset for consistency and reliability, we proceed by utilizing advanced DL models for automated classification. Pre-trained models known for their robustness in image analysis are employed to improve classification accuracy and efficiency, reducing the need for extensive new model training. The ResNet family is among the most popular architectures for computer vision tasks, leveraging residual connections to mitigate the vanishing gradient problem and enabling the training of deep networks. ResNet18, with its eighteen layers, is relatively shallow but still performs well across various classification tasks. As we increase the network depth, such as in ResNet34, performance improves, offering greater capacity for learning complex features, though at the cost of higher computational requirements \cite{he2016deep}.

We employ EfficientNet and ConvNeXt as base architectures due to their strong performance in visual tasks. ConvNeXt integrates transformer-inspired enhancements while retaining the efficiency of CNNs \cite{liu2022convnet}. In particular, ConvNeXt+Attn uses grouped convolutions and localized attention to emphasize relevant patterns, outperforming ViT-based models in this domain.

We are aiming to train a model on expert-cropped images, which we refer to as the teacher model. This model is specifically designed to focus on regions that are deemed most relevant for classification, as determined by expert knowledge. By training the teacher model exclusively on these cropped regions, we eliminate unnecessary background noise and provide a more structured learning framework. The key motivation behind this approach is to develop a high-confidence feature extraction model that relies on explicit human-guided supervision to achieve superior classification accuracy.

However, while the teacher model benefits from such guided supervision, it inherently lacks the ability to generalize to full images where the location of discriminative regions is unknown. In real-world applications, we often do not have access to precise region annotations, and cropping images manually is both labor-intensive and impractical at scale. This constraint leads us to the necessity of training a student model, which learns from full, unaltered images without relying on explicit cropping. 

The goal of the student model is twofold: (1) to learn the important regions autonomously without requiring external guidance, and (2) to achieve closer results to the teacher model, despite receiving full, unprocessed input data. Instead of restricting the training to predefined regions, we allow the student model to extract relevant information dynamically, simulating a real-world scenario where attention needs to be self-discovered.


\subsection{Knowledge Distillation for Spatial Logits Learning}

In this work, we propose a novel approach to training deep learning models for dental image classification by leveraging knowledge distillation to transfer spatial information from a teacher model—trained on cropped images—to a student model—trained on full, uncropped images. The primary objective is to enable the student model to learn the spatial distribution of logits from the teacher model without the necessity of explicit cropping steps during inference. This eliminates the dependency on an external object detection network (e.g., YOLO) and ensures that the model remains robust to variations in image framing. Furthermore, the incorporation of spatial knowledge transfer enhances the generalization capabilities of the student model, allowing it to classify images with varying degrees of background information while preserving the discriminative spatial characteristics learned from the teacher model.

\subsection{Training Framework}
The training process consists of two models sharing the same architecture: a teacher model and a student model. The teacher model is trained on cropped dental images, where the regions of interest (ROI) are explicitly extracted. These cropped images allow the model to learn spatially localized features that are highly relevant to classification. In contrast, the student model is trained on uncropped images, which contain additional background information that could potentially introduce noise. To ensure that the student model effectively mimics the behavior of the teacher, we introduce a knowledge distillation framework where the student learns to generate spatial logits similar to those of the teacher, despite operating on full images. 
\begin{figure*}[htbp]
\centering
\includegraphics[width=0.9\textwidth]{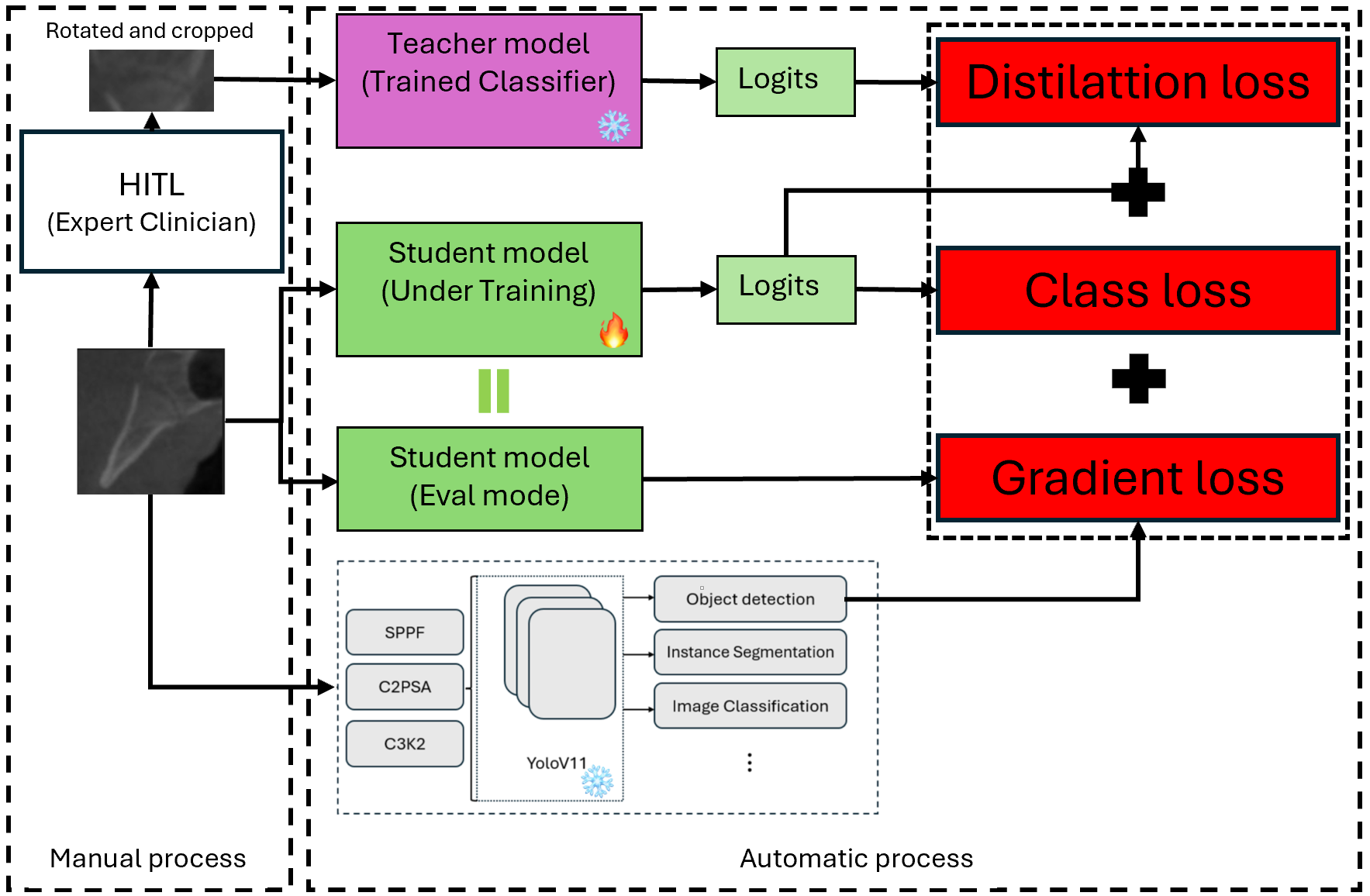}
\caption{Knowledge Distillation and YOLO-Based Automation with Gradient Loss for SOS Fusion Staging}
\label{fig:Orientation_and_Crop}
\end{figure*} 
The student model learns using a total loss function, which is a weighted combination of attention loss, Distillation loss, and classification loss, as defined in Eq.~(\ref{eq:expanded_total_loss}). The function is carefully designed to balance the trade-off between spatial alignment, classification accuracy, and knowledge transfer.

\subsection{Loss Functions for Student Model Training}
The student model is trained using a combination of three loss functions:
distillation loss, gradient-based attention loss, and classification loss. Each loss function plays a distinct role in ensuring that the student model retains both spatial understanding and classification accuracy. The total loss function is mathematically expressed as:

\begin{equation}
\begin{aligned}
\mathcal{L}_{total} = & \, \alpha \,\frac{1}{N} \sum_{i=1}^{N} \Bigl(\widehat{A}_i - M_i \Bigr)^2 \\
& + \beta T^2 \sum_{k=1}^{K} P_k \left( \log P_k - \log Q_k \right) \\
& + \theta \Bigl(- \sum_{k=1}^{K} y_k \,\log(\hat{y}_k)\Bigr)
\end{aligned}
\label{eq:expanded_total_loss}
\end{equation}

where \( \mathcal{L}_{attn} \) represents the spatial attention loss, \( \mathcal{L}_{dist} \) is the temperature-scaled distillation loss, and \( \mathcal{L}_{cls} \) is the classification loss formulated as the cross-entropy between predicted and true labels. The hyperparameters \( \alpha, \beta, \theta \) control the relative contribution of each term to the total loss.

For knowledge distillation, we use a temperature parameter \( T = 3 \) to smooth the probability distributions of the teacher and student models, ensuring better generalization and gradient flow. The KL divergence term is scaled by \( T^2 \) to maintain proper gradient magnitudes, following the formulation introduced in \cite{hinton2015distilling}.

Through extensive experimentation, the best results were achieved with a weighting of \( \alpha = 0.01 \), \( \beta = 0.8 \), and \( \theta = 0.2 \), ensuring a balance between preserving spatial information and optimizing classification performance.

\subsection{Distillation Loss for Spatial Logits Alignment}  

Since the teacher model operates on cropped images, its logits focus on key regions, learning representations that are highly concentrated within the ROI. To transfer this knowledge, we minimize the divergence between the teacher's and student's spatial logit distributions. The objective is to allow the student to adapt to full images while maintaining attention on the most important anatomical structures. 

The distillation loss is formulated as Kullback-Leibler (KL) divergence:

\begin{equation}
\label{eq:distillation_loss}
\mathcal{L}_{dist} \;=\; T^2 \sum_{k=1}^{K} P_k \left( \log P_k - \log Q_k \right),
\end{equation}

where \( P_k = \mathbf{z}_{\text{teacher}}(k) \) and \( Q_k = \mathbf{z}_{\text{student}}(k) \) denote the probability distributions of the teacher and student models, respectively. These distributions are often smoothed using a temperature parameter \( T \), which controls the softening of logits during knowledge transfer:

\begin{equation}
P_k = \frac{\exp(\mathbf{z}_{\text{teacher}}(k) / T)}{\sum_{j=1}^{K} \exp(\mathbf{z}_{\text{teacher}}(j) / T)}
\end{equation}

\begin{equation}
Q_k = \frac{\exp(\mathbf{z}_{\text{student}}(k) / T)}{\sum_{j=1}^{K} \exp(\mathbf{z}_{\text{student}}(j) / T)}
\end{equation}

where increasing \( T \) results in softer probability distributions, facilitating smoother knowledge transfer. 
For our experiments, we set \( T = 3 \) to balance knowledge transfer smoothness and gradient stability.
The KL loss is scaled by \( T^2 \) to ensure that gradients remain appropriately scaled during backpropagation.

\subsection{Gradient-Based Attention Loss (Grad-CAM Loss)}
To refine the spatial focus of the student model, we introduce a gradient-based penalty derived from Gradient-weighted Class Activation Mapping (Grad-CAM). Grad-CAM generates class-specific heatmaps that highlight the most influential regions for a given classification prediction. However, without explicit guidance, the student model may focus on irrelevant regions. To ensure that the student model aligns its attention with the anatomically relevant regions, we compare its Grad-CAM activation maps with the YOLO-derived region of interest (ROI), rather than the teacher’s attention maps.

During training, the YOLO-based object detection model is used to provide a ground-truth attention map \( M_i \), which represents the critical spatial areas that should contribute to classification. The student model’s Grad-CAM heatmap \( \widehat{A}_i \) should ideally align with this reference. We enforce this alignment through an attention loss function, defined as:

\begin{equation}
\mathcal{L}_{attn} = \frac{1}{N} \sum_{i=1}^{N} \Bigl(\widehat{A}_i - M_i \Bigr)^2,
\label{eq:attention_loss}
\end{equation}

where \( \widehat{A}_i \) represents the student model's Grad-CAM heatmap and \( M_i \) is the YOLO-generated attention mask. This term penalizes discrepancies between the predicted and expected attention, forcing the student model to focus on the same anatomical regions identified by the YOLO detector.

\subsubsection{Gradient Optimization for Spatial Alignment}  

To better understand how the student model adjusts its attention to match the YOLO-derived focus areas, we analyze the gradient of the attention loss function:

\begin{equation}
\frac{\partial \mathcal{L}_{attn}}{\partial \widehat{A}_i} = \frac{2}{N} \left( \widehat{A}_i - M_i \right).
\end{equation}

This gradient formulation highlights that when the student’s Grad-CAM heatmap aligns well with the YOLO attention mask, the gradient approaches zero, requiring minimal updates. Conversely, large deviations result in strong correction signals, forcing the student model to adapt its focus iteratively over training epochs.

Since Grad-CAM maps are derived from intermediate convolutional feature activations, we formally express the student’s Grad-CAM activation map as:

\begin{equation}
\widehat{A}_i = \sum_{c} w_c A_i^c,
\end{equation}

where \( A_i^c \) represents the activation of channel \( c \) at spatial location \( i \), and \( w_c \) is a weighting factor computed via global average pooling over the gradients:

\begin{equation}
w_c = \frac{1}{Z} \sum_{i} \frac{\partial \mathcal{L}_{cls}}{\partial A_i^c}.
\end{equation}

This formulation explicitly enforces that the attention mechanism is optimized based on class-specific features, ensuring that the student model prioritizes the same anatomical regions identified by YOLO without requiring explicit segmentation during inference.

By enforcing this attention-based constraint, we eliminate the need for a separate cropping step at inference time, allowing the student model to autonomously focus on the correct regions without external object detection models.

\subsection{Classification Loss}
A standard classification loss, formulated as cross-entropy loss, is applied to ensure that the student model correctly predicts class labels based on ground truth annotations. This loss plays a crucial role in maintaining the classification accuracy of the student model while benefiting from the distillation and Grad-CAM-based spatial alignment mechanisms.

\begin{equation}
\mathcal{L}_{cls} = - \sum_{k=1}^{K} y_k \,\log(\hat{y}_k)
\label{eq:classification_loss}
\end{equation}

where \( y_k \) represents the ground-truth label distribution and \( \hat{y}_k \) is the predicted probability distribution.

One of the key advantages of our approach is that it removes the need for an external object detection model (e.g., YOLO) to crop the region of interest before classification during inference. 

A two-stage approach is commonly employed, where YOLO first detects the ROI, followed by a classification model processing the cropped region. This method achieves an improved accuracy of 82.5\%, which is higher than the 79.97\% accuracy obtained when training and classifying directly on full images. This suggests that segmentation helps the model focus on critical structures, reducing background interference. However, the two-stage approach introduces potential inaccuracies due to incorrect cropping or the loss of contextual information outside the detected ROI. Additionally, this pipeline requires additional computational overhead, making inference less efficient compared to an end-to-end classification model that learns spatial attention implicitly.

By training the student model to learn spatial attention implicitly through knowledge distillation and Grad-CAM-based supervision, we eliminate the necessity for a separate object detection step. Instead, our framework ensures that the student model learns to focus on the correct anatomical regions within the full image, enabling a single, end-to-end classification model that is both efficient and robust to spatial variations.

\bigskip

\noindent In summary, our proposed knowledge distillation framework allows the student model to successfully learn spatially meaningful representations from the teacher model without explicit cropping, thereby enhancing classification performance while maintaining robustness to image framing variations.

\begin{table}[t]
\centering
\small
\renewcommand{\arraystretch}{1.1}
\caption{Performance of Models Trained and Tested on Expert-Cropped Images (Performance averaged over 5-fold cross-validation)}

\label{tab:teacher_results}
\begin{tabular}{lcccc}
\toprule
Model & Acc (\%) & Prec (\%) & Rec (\%) & F1 (\%) \\
\midrule
EfficientNet\_b0   & 78.57 & 78.48 & 78.57 & 77.73 \\
ResNet18           & 84.87 & 84.51 & 84.87 & 84.34 \\
ResNet34           & 84.59 & 84.40 & 84.59 & 84.14 \\
ResNet50           & 84.03 & 83.95 & 84.03 & 83.33 \\
ConvNeXt           & 87.39 & 87.48 & 87.39 & 87.24 \\
\textbf{ConvNeXt+Attn} & \textbf{88.24} & \textbf{88.93} & \textbf{88.24} & \textbf{88.09} \\
\bottomrule
\end{tabular}
\end{table}


\begin{table*}[tp]
    \centering
\caption{Performance of Models Trained and Tested on Full Images With and Without the Proposed Framework (Performance averaged over 5-fold cross-validation)}

    \label{tab:classification_results}
    \resizebox{\textwidth}{!}{%
    \begin{tabular}{llcccc}
        \toprule
        Model & Version & Accuracy (\%) & Precision (\%) & Recall (\%) & F1 (\%) \\
        \midrule
        EfficientNet-b0 & Baseline & 70.58  & 69.52  & 70.58  & 69.13 \\
        ResNet18         & Baseline & 72.55  & 72.12  & 72.55  & 70.84 \\
        ResNet34         & Baseline & 75.07  & 75.01  & 75.07  & 74.05 \\
        ResNet50         & Baseline & 74.79  & 74.08  & 74.79  & 73.79 \\
        ViT\_Large       & Baseline & 64.42  & 63.40  & 64.42  & 63.03 \\
        ConvNeXt         & Baseline & 78.99  & 78.57  & 78.99  & 78.35 \\
        \midrule
        ConvNeXt + Attn           & Baseline & 79.97  & 80.65  & 79.97  & 78.87 \\
        ConvNeXt + Attn           & \textbf{Proposed Framework} & \textbf{83.75}  & \textbf{84.01}  & \textbf{83.75}  & \textbf{83.29} \\
        \bottomrule
    \end{tabular}%
    }
\end{table*}


\section{Experimental Validation}
To validate the effectiveness of our method, we first evaluate the performance of the teacher model, which is trained and tested on expert-cropped images. Since these images are carefully selected by experts, the model benefits from a highly focused learning process. Table~\ref{tab:teacher_results} presents the classification performance of the teacher model under these conditions. This serves as the upper bound for performance, as the model has direct access to the most relevant regions of the input.

Before evaluating classification performance, we first assess the object detection capabilities of YOLOv11, which serves as a reference for guiding the model’s spatial attention during training. YOLOv11 achieves a mAP@0.5 of 72.8\%, indicating strong detection capabilities across different anatomical regions. While our framework does not rely on explicit object detection, this result highlights the model’s ability to recognize key regions effectively. By integrating Gradient-Based Attention Loss (Grad-CAM Loss) and leveraging YOLO’s detection signal implicitly, we enable the student model to focus on critical structures without the need for pre-cropping.

Next, we evaluate the performance of models trained and tested on full images, including our proposed student model in the New Framework. Unlike the teacher model, which benefits from expert cropping, these models must learn to identify and focus on relevant regions without explicit spatial guidance. However, our proposed framework, which incorporates knowledge distillation and spatial logits learning, enables the student model to achieve results significantly closer to the teacher model. Table~\ref{tab:classification_results} presents a comparison of different architectures trained on full images.

The results indicate that the New Framework, built upon the ConvNeXt architecture, achieves the highest classification performance among models trained on full images. Specifically, the New Framework attains an accuracy of 83.75\%, which is 3.78\% higher than the baseline ConvNeXt model (79.97\%) and 8.68\% higher than ResNet34 (75.07\%). While still slightly below the teacher model’s performance, this demonstrates the effectiveness of our proposed training methodology in narrowing the gap.

Furthermore, the Precision (84.01\%) and Recall (83.75\%) of the New Framework surpass all other models in this category, reinforcing its ability to balance predictions across different classes. The key advantage of the New Framework lies in its integration of knowledge distillation and spatial logits learning, which significantly enhance the student model’s ability to focus on relevant features despite being trained on full images.

\begin{figure}[t]
    \centering
    \begin{minipage}{0.48\linewidth}
        \centering
        \includegraphics[width=0.9\linewidth]{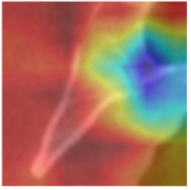}
        \caption*{(a) Without framework}
    \end{minipage}\hfill
    \begin{minipage}{0.48\linewidth}
        \centering
        \includegraphics[width=0.9\linewidth]{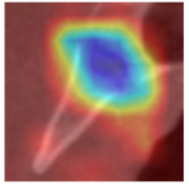}
        \caption*{(b) Proposed Framework}
    \end{minipage}
    \caption{Comparison of Grad-CAM heatmaps generated by the ConvNeXt + Attn model with (b) and without (a) the proposed framework. The proposed approach yields more localized and clinically relevant attention.}
    \label{fig:heatmap_comparison}
\end{figure}

Beyond these quantitative improvements, it is worth noting that the New Framework consistently outperforms vision transformers such as ViT\_Large, which achieves only 64.42\% accuracy. This suggests that transformer-based models may struggle to localize critical anatomical features in full images without explicit spatial priors. In contrast, convolutional architectures with attention mechanisms, such as ConvNeXt+Attn, demonstrate superior generalization in this setting. The consistent margin of improvement across multiple metrics also underscores the robustness of our approach, highlighting its potential applicability beyond the current dataset. The ability to bridge the performance gap between full-image and expert-cropped settings is crucial for real-world deployment, where manual pre-processing may not be feasible. By leveraging spatial attention and knowledge transfer, our framework offers a practical and scalable solution for medical image classification tasks. To further illustrate the effectiveness of the proposed framework, we visualize the attention heatmaps generated by the ConvNeXt + Attn model with and without our framework. As shown in Fig.~\ref{fig:heatmap_comparison}, the proposed method leads to more focused and anatomically relevant attention regions on the synchondrosis.

\section{Discussion}
\label{sec:discussion}
Our findings confirm that combining knowledge distillation with attention-based losses leads to notable performance gains for fully automated SOS fusion staging. By training a “teacher” on manually cropped images and transferring this localized spatial focus to a “student” that processes uncropped images, we effectively bypass the need for external segmentation tools. This simplified pipeline lessens computational overhead and potential errors introduced by ROI detection modules, making large-scale clinical deployment more feasible.

The combination of distillation and Grad-CAM-based attention improves both accuracy and interpretability by aligning student model focus with clinically meaningful regions.

Despite promising results, several challenges remain. Our approach still relies on manual cropping during the teacher’s training phase, demanding some level of expert input. Future research could explore weakly supervised or fully automated cropping strategies to further streamline training. Additionally, larger and more diverse datasets would validate the model’s robustness across varied imaging conditions and patient demographics. Nonetheless, the proposed framework shows strong potential for improving diagnostic accuracy and reducing the overall complexity of clinical workflows, paving the way for broader adoption in orthodontics, forensic anthropology, and other medical imaging domains.
\section{Conclusion}

We presented an interpretable deep learning framework for automated SOS fusion staging that bypasses the need for external segmentation models during inference. Our training pipeline utilizes YOLO-based cropping to guide the model in learning relevant spatial attention. Once trained, however, the student model alone processes uncropped data, reducing system complexity, minimizing the overall parameter count, and accelerating inference. By obviating the need for additional segmentation modules at deployment, the proposed method offers a more streamlined and practical solution for clinical and forensic settings.
 
Future work will focus on further automating the data preparation phase to reduce manual supervision, expanding our dataset to encompass a broader range of demographic and clinical variations, and improving the generalizability of our approach. Beyond SOS fusion staging, we envision the proposed architecture and knowledge distillation strategy being extended to other medical imaging tasks requiring localized feature extraction and high interpretability.

\label{sec:conclusion}

\label{sec:typestyle}
\clearpage


\end{document}